\documentclass[conference]{IEEEtran}
\IEEEoverridecommandlockouts

\usepackage{float}
\usepackage{booktabs}
\usepackage{amssymb}
\usepackage{subcaption}
\usepackage[font=small,labelsep=period]{caption} 
\usepackage{cite}
\usepackage{amsmath,amssymb,amsfonts}
\usepackage{hyperref}
\usepackage{graphicx} 
\graphicspath{ {image/} }  
\usepackage{float}
\usepackage{caption}
\hypersetup{hidelinks,
    colorlinks=true,
    allcolors=black,
    pdfstartview=Fit,
    breaklinks=true}
\usepackage{algorithmic}
\usepackage{textcomp}
\usepackage{xcolor}
\usepackage{times}
\usepackage{mathptmx}
\def\BibTeX{{\rm B\kern-.05em{\sc i\kern-.025em b}\kern-.08em
    T\kern-.1667em\lower.7ex\hbox{E}\kern-.125emX}}
    
\begin{document}

\title{HMPNet: A Feature Aggregation Architecture for Maritime Object Detection from a Shipborne Perspective
\thanks{This work was supported by the National Natural Science 
Foundation of China (Grant No. 52401362), National Training 
Program of Innovation and Entrepreneurship for Undergraduates
(202410057072).}
}

\author{
\IEEEauthorblockN{1\textsuperscript{st} Yu Zhang\IEEEauthorrefmark{1}}
\IEEEauthorblockA{
Tianjin University of Science and Technology\\
Tianjin, China\\
\texttt{zhangyuai@tust.edu.cn}}
\and
\IEEEauthorblockN{2\textsuperscript{nd} Fengyuan Liu\IEEEauthorrefmark{1}}
\IEEEauthorblockA{
Tianjin University of Science and Technology\\
Tianjin, China\\
\texttt{fengyuan711@gmail.com}}
\and
\IEEEauthorblockN{3\textsuperscript{rd} Juan Lyu\IEEEauthorrefmark{2}}
\IEEEauthorblockA{
Tianjin University of Science and Technology\\
Tianjin, China\\
\texttt{lvjuan@tust.edu.cn}}
\and
\IEEEauthorblockN{4\textsuperscript{th} Yi Wei}
\IEEEauthorblockA{
Tianjin Institute of Navigation Instruments\\
Tianjin, China\\
\texttt{hityiwei@163.com}}
\and
\IEEEauthorblockN{5\textsuperscript{th} Changdong Yu}
\IEEEauthorblockA{
Dalian Maritime University\\
Dalian, China\\
\texttt{ycd@dlmu.edu.cn}}
\thanks{\IEEEauthorrefmark{1}These authors contributed equally to this work.}
\thanks{\IEEEauthorrefmark{2}Corresponding author.}
}

\maketitle
\begin{abstract}
In the realm of intelligent maritime navigation, object detection from a shipborne perspective is paramount. Despite the criticality, the paucity of maritime-specific data impedes the deployment of sophisticated visual perception techniques, akin to those utilized in autonomous vehicular systems, within the maritime context. To bridge this gap, we introduce Navigation12, a novel dataset annotated for 12 object categories under diverse maritime environments and weather conditions. Based upon this dataset, we propose HMPNet, a lightweight architecture tailored for shipborne object detection. HMPNet incorporates a hierarchical dynamic modulation backbone to bolster feature aggregation and expression, complemented by a matrix cascading poly-scale neck and a polymerization weight sharing detector, facilitating efficient multi-scale feature aggregation. Empirical evaluations indicate that HMPNet surpasses current state-of-the-art methods in terms of both accuracy and computational efficiency, realizing a 3.3$\%$ improvement in mean Average Precision over YOLOv11n, the prevailing model, and reducing parameters by 23$\%$. Code is available at: \url{https://github.com/tustAilab/HMPNet}
\end{abstract}

\section{Introduction}
\label{sec:intro}

Maritime object detection is essential for route planning, collision avoidance \cite{r1}, and intelligent shipping, serving as a critical technology to ensure navigation safety and improve shipping efficiency \cite{rE}. This technology enables real-time identification of ships, buoys, lighthouses, and other maritime objects, thereby optimizing route planning and reducing collision risks. Furthermore, it accelerates target localization in search and rescue operations \cite{r2} and has substantial implications for ocean monitoring and maritime management.

Although there has been some research on maritime object detection, such as the anchor-free detection model LPEDet proposed by Feng et al. \cite{r4}, the GL-DETR model based on local multi-scale features by Li et al. \cite{r5}, the ship detection network BLNet based on balanced learning by Zhang et al. \cite{r6}, and the CHEANet model based on contour keypoint detection proposed by Gao et al. \cite{r7}, these studies mainly focus on remote sensing (RS) or synthetic aperture radar (SAR) imagery, which fail to reflect the complex and dynamic scenarios encountered in real-world ship navigation \cite{r9}. Remote sensing images typically offer a large field of view and high global resolution but lack detailed features. Similarly, SAR images perform well under low-light conditions, however, they generally suffer from low resolution and limited texture details, making it challenging to accurately distinguish multi-class and multi-scale objects \cite{r10}. In contrast, maritime images captured from shipborne perspectives better reflect the complex and dynamic scenarios of real-world ship navigation, thus facilitating intelligent visual perception and real-time decision-making. However, tasks from shipborne perspectives face significant challenges, including vast depth-of-field ranges that lead to considerable variations in object scales and highly dynamic backgrounds influenced by factors such as waves and weather conditions \cite{r11}, which impose stricter requirements on the accuracy and generalization ability of detection models.
\begin{figure*}[!t]
  \centering
  \includegraphics[width=0.6\textwidth]{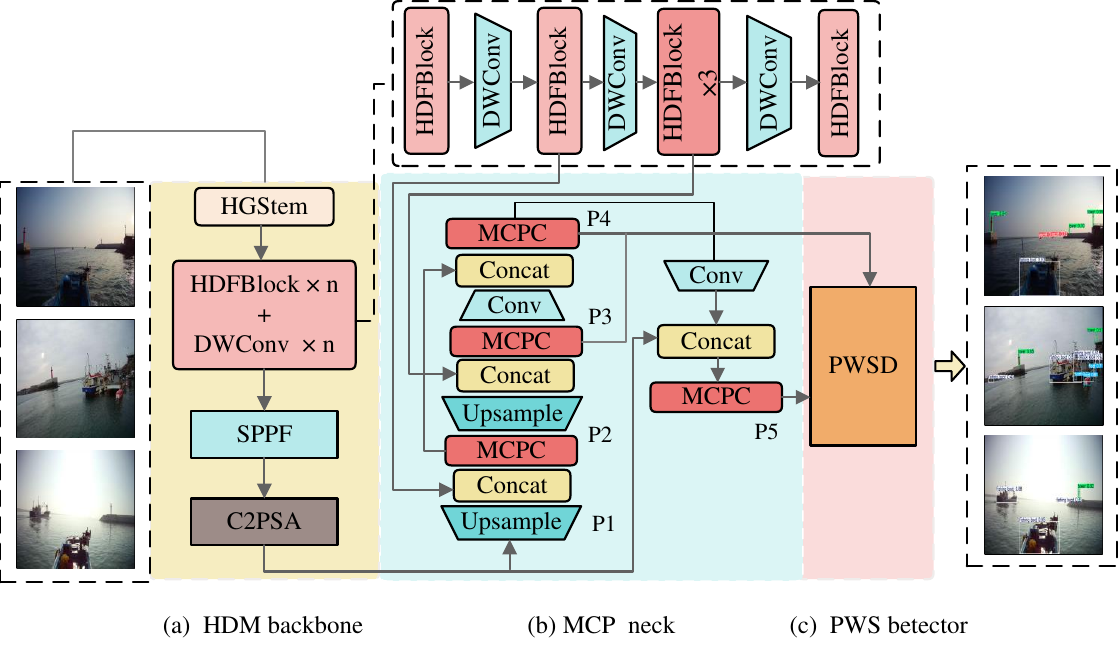}
  \caption{The architecture of the proposed HMPNet consists of three components, including (a) the HDM backbone, (b) the MCP neck, and (c) the PWS detector.}
  \label{fig:1}
\end{figure*}
Currently, object detection methods for general scenarios have been widely applied with remarkable success in fields such as autonomous driving, facial recognition, and so on, offering valuable insights for the design of shipborne object detection approaches. For example, the two-stage Faster R-CNN \cite{r12} achieves high-precision object localization and classification through a region proposal network; SSD \cite{r13} enhances multi-object detection capabilities using a multi-scale feature pyramid; the YOLO series \cite{r14,r15,r16,r17,r18} significantly improves detection speed with its single-stage framework; and RT-DETR \cite{r19}, leveraging a self-attention mechanism, captures global features and demonstrates considerable potential in complex scenarios. However, these methods typically involve high parameter counts and computational complexity, making them less ideal for direct application to shipborne tasks that require lightweight and real-time performance. These limitations become particularly evident when addressing significant scale variations and highly dynamic maritime backgrounds, challenges that still demand further investigation and optimization.

To address these issues, we constructed Navigation12, the first high-quality shipborne object detection dataset, encompassing 12 target categories across diverse maritime regions, scenarios, and weather conditions. These categories include a wide range of objects such as ships, buoys, lighthouses, reefs, wind turbines, etc. Based on this dataset, we propose HMPNet, a lightweight multi-scale feature aggregation architecture that integrates a hierarchical dynamic modulation (HDM) backbone, a matrix cascading poly-scale (MCP) neck, and a polymerization weight sharing (PWS) detector. Through dynamic feature modulation, multi-scale aggregation, and a lightweight design, HMPNet effectively addresses multi-scale, multi-class, and complex-background challenges in shipborne scenarios, substantially improving detection accuracy and efficiency.

The main contributions of this paper are as follows:
\begin{itemize}
  \item We develop Navigation12, the first shipborne object detection dataset comprising over 18,000 high-resolution maritime images with annotations for 12 target categories across diverse maritime regions, scenarios, and weather conditions. To the best of our knowledge, it is currently the largest and most diverse dataset dedicated to maritime shipborne object detection.
  \item We introduce HMPNet, which achieves efficient feature representation with a HDM backbone, enhances feature interaction and aggregation through the MCP neck, and improves feature fusion efficiency while reducing parameters with a PWS detector. This architecture achieves an effective balance between detection accuracy and computational cost.
  \item We comprehensively evaluate the detection performance of HMPNet on the Navigation12 dataset. Experimental results demonstrate that HMPNet achieves a remarkable 80.9$\%$ mAP with only 2M parameters, significantly outperforming various state-of-the-art detection methods.
\end{itemize}

\section{Methodology}

To effectively address the challenges of object scale variation and complex backgrounds in shipborne perspectives, this paper proposes a lightweight feature aggregation architecture, HMPNet. Fig. \ref{fig:1} illustrates the overall architecture of HMPNet, which consists of three components: the backbone network (HDM backbone), the feature aggregation module (MCP neck), and the detection head (PWS detector). The HDM backbone primarily performs feature extraction and aggregates shipborne image features. In the MCP neck, multi-level features extracted from the backbone network undergo deep interaction and aggregation, enhancing the representation of these features. Finally, the PWS detector receives the aggregated features, optimizing feature reuse and reducing computational costs by sharing convolutional weights, and then outputs the detection results. The design of HMPNet fully embodies the concept of feature aggregation, effectively addressing challenges related to object scale variations and complex backgrounds, thereby providing robust support for intelligent maritime navigation.

\begin{figure}[!t]
  \centering
  \includegraphics[width=8cm,page={1}]{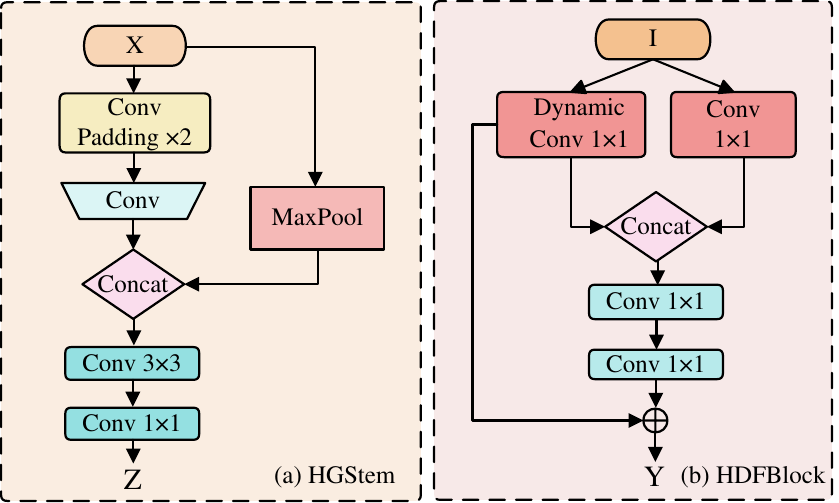}
  \caption{The structures of HGStem and HDFBlock. (a) HGStem, utilizes multi-scale convolution and pooling operations to extract initial features. (b) HDFBlock, performs hierarchical dynamic feature aggregation through dynamic convolutions and depthwise separable convolutions.}
  \label{fig:2}
\end{figure}

\subsection{HDM Backbone}
The backbone network of the HMPNet architecture focuses on extracting fundamental features from shipborne perspective images and conducting initial fusion processing. As shown in Fig. \ref{fig:1}(a), our HDM backbone first employs the HGStem module \cite{r20} as the initial aggregation unit for input features. Next, we introduce a hierarchical dynamic focus block (HDFBlock), which dynamically adapts to target features of varying scales and positions through a hierarchical focusing strategy. This block leverages depthwise separable convolutions (DWConv) \cite{r21} to iteratively extract and aggregate multi-scale features. Finally, initial feature fusion is achieved using the Spatial Pyramid Pooling – Fast (SPPF) \cite{r18} and Convolutional Block with Parallel Spatial Attention (C2PSA) modules \cite{r18}, enhancing the capability to capture global information.

The HGStem efficiently extracts initial features through multi-level convolution and pooling operations, providing refined feature representations for subsequent modules, as illustrated in Fig. \ref{fig:2} (a). Subsequently, as shown in Fig. \ref{fig:2} (b), the HDFBlock processes the features extracted by HGStem and branches them into dynamic convolution \cite{r22} and pointwise convolution operations. This design aims to enhance feature focusing capability and ensure computational efficiency while improving feature aggregation. The features are then concatenated and refined through two consecutive pointwise convolutions for layer-wise extraction and nonlinear mapping, enhancing the representation capacity of the features. Finally, the output is connected to the input features via residual connections to ensure efficient gradient aggregation. With the hierarchical dynamic focusing strategy, HDFBlock can effectively strengthen the representation of targets at different scales, meeting the high-precision feature representation requirements for target detection tasks in complex maritime scenarios. 

\begin{figure}[!t]
  \centering
  \includegraphics[width=9cm,page={1}]{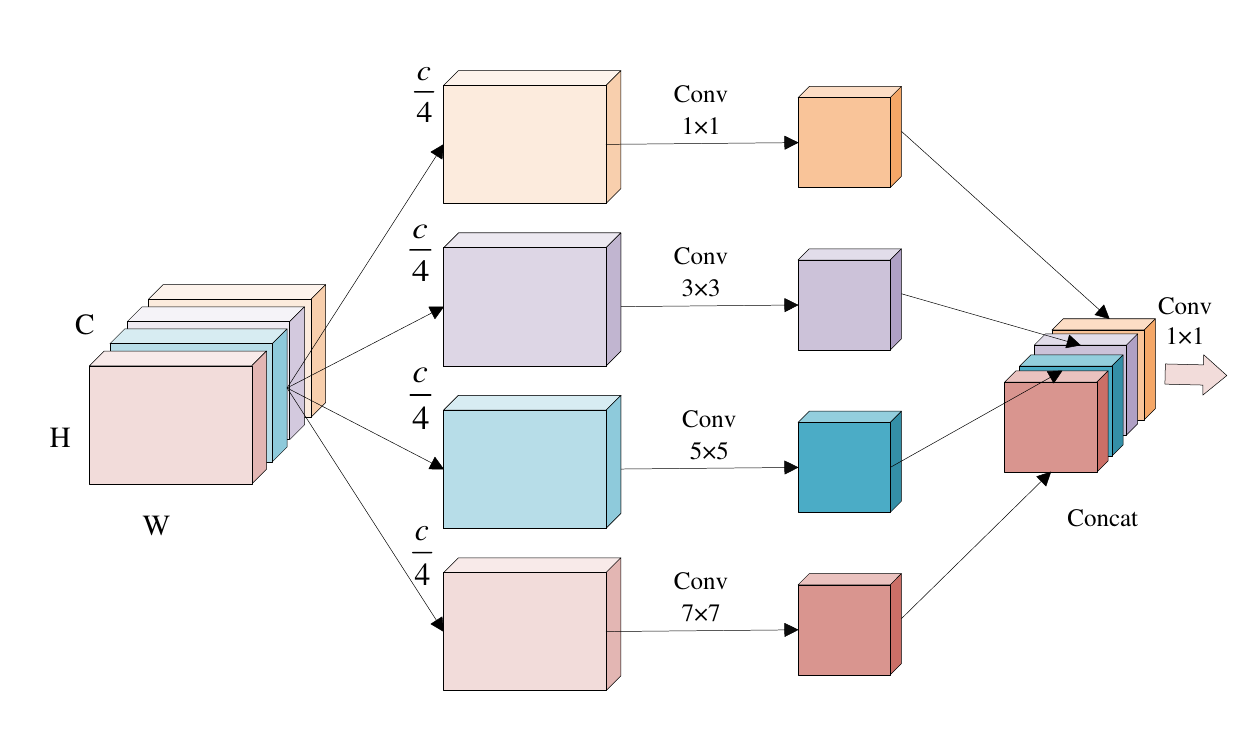}
  \caption{Structure of MCPC module.}
  \label{fig:3}
\end{figure}

\subsection{MCP Neck}
In HMPNet, the MCP neck acts as a feature fusion module, aggregating both high-semantic deep features and low-semantic detail features from the backbone network. This process enables the integration of local information with global contextual information, achieving unified modeling. As shown in the MCP neck section of Fig. \ref{fig:1} (b), the deep features, upsampled and restored, are first concatenated with shallow features to enhance the interaction between local details and contextual information. Next, to further strengthen multi-scale feature aggregation, we design the MCPC module based on the efficient feature grouping ideas from Shufflenet \cite{r23} and SMT \cite{r24}. This module performs grouped convolution on the fused features, ensuring consistent feature representation across targets of different scales. The structure of the MCPC module is shown in Fig. \ref{fig:3}. Unlike Shufflenet and SMT , the MCPC module divides the input features into equally sized groups and processes each group in parallel using convolutions with different kernel sizes, extracting features at different receptive fields and performing cross-scale concatenation. The average grouping strategy balances the contribution of each group to the feature representation, preventing any single feature from dominating the fusion process, thereby capturing multi-scale information more effectively. Finally, global contextual information is aggregated through 1 × 1 convolutions. This design significantly reduces computational load while preserving rich detail information.

Through the above feature fusion process, MCP neck is able to significantly enhance the ability of HMPNet to aggregate features for targets of different scales, while also improving real-time processing and overall generalization capabilities.

\begin{figure}[!t]
  \centering
  \includegraphics[width=8.5cm,page={1}]{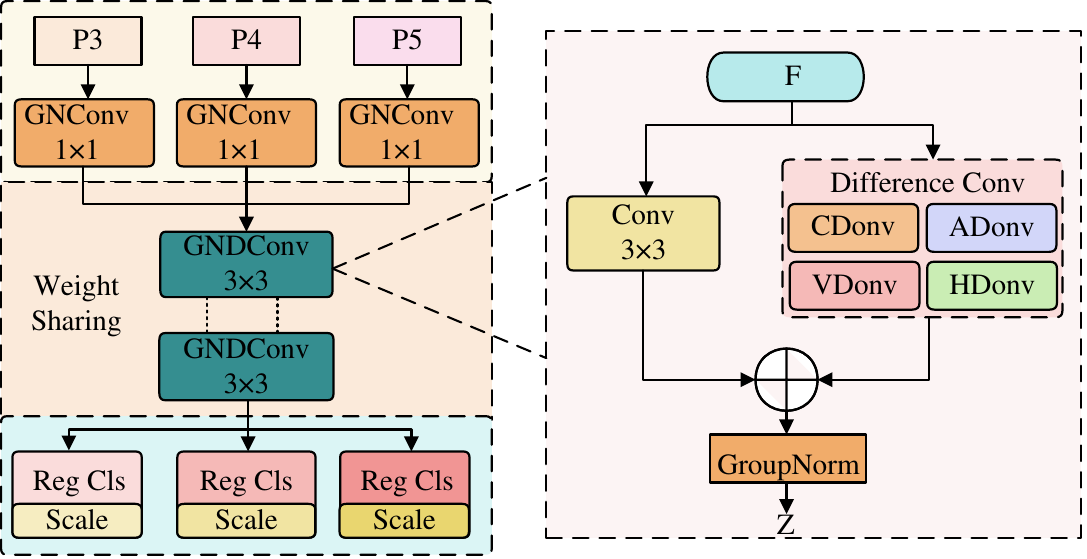}
  \caption{Structure of PWS detector. The PWS detector reduces the number of parameters through shared convolution, incorporates GNDConv to enhance detail capture, and utilizes the Scale layer for adaptive adjustment of multi-scale features, ensuring detection accuracy and robustness.}
  \label{fig:4}
\end{figure}

\subsection{PWS Detector}
In HMPNet, the PWS detector achieves target classification and bounding box regression utilizing the features aggregated from different scales in the MCP neck. As shown in Fig. \ref{fig:4}, the PWS detector first applies group-normalized convolution (GNConv) to the features from the three stages (P3, P4, and P5) of the MCP neck, as depicted in Fig. \ref{fig:1} (b), performing initial channel compression. These features then undergo detail enhancement and normalization through GNDConv. The GNDConv layers are connected through a weight-sharing mechanism to reduce computational costs. Finally, to address scale differences across detection layers, a scale layer first performs feature scaling, after which losses are computed through the Reg/Cls branches at corresponding scales. 

GNDConv, the pivotal element within the PWS detector framework, draws inspiration from the DEConv module employed in the DEA-Net architecture, as delineated in prior work  \cite{r25}. This module is used to augment the feature representation and generalization capabilities of the model by integrating prior information. Unlike DEConv, we introduce the polymerization weight sharing mechanism, which enhances multi-scale feature aggregation by sharing convolutional weights across detection layers. This strategy ensures consistent feature extraction across scales while reducing redundancy, effectively avoiding additional parameters and computational costs. Meanwhile, GNDConv uses group normalization (GN) \cite{r26} instead of the traditional Batch Normalization (BN), which normalizes features based on channel grouping. This approach eliminates the dependence on global statistics from batch sizes, allowing the network to more stably capture the detailed features of the target.

As illustrated in Fig. \ref{fig:4}, in the structure of GNDConv, various types of differential convolutions are used to integrate prior information, enhancing the detail-capturing capability of convolution operations. For an input feature $F$, the output of GNDConv $N$, is expressed as
\begin{equation}
N=\mathrm{GroupNorm}\left(\mathrm{Conv}\left(F\right)+\sum_{i=1}^{n}{\mathrm{DiffConv}}_i\left(F\right)\right), \label{con:inventoryflow}
\end{equation}
where DiffConv$_i$ represents different types of detail-enhancing convolutions, which perform differential processing on the input features based on various priors, such as direction and magnitude. Additionally, by employing the weight-sharing mechanism, the PWS detector significantly reduces the number of parameters, making the detector more lightweight while maintaining high sensitivity to details. This approach effectively addresses the feature distribution variations caused by uneven target scales and density distributions in maritime scenarios.

\section{Experiment}
\begin{figure*}
\centering
\includegraphics[width=\textwidth]{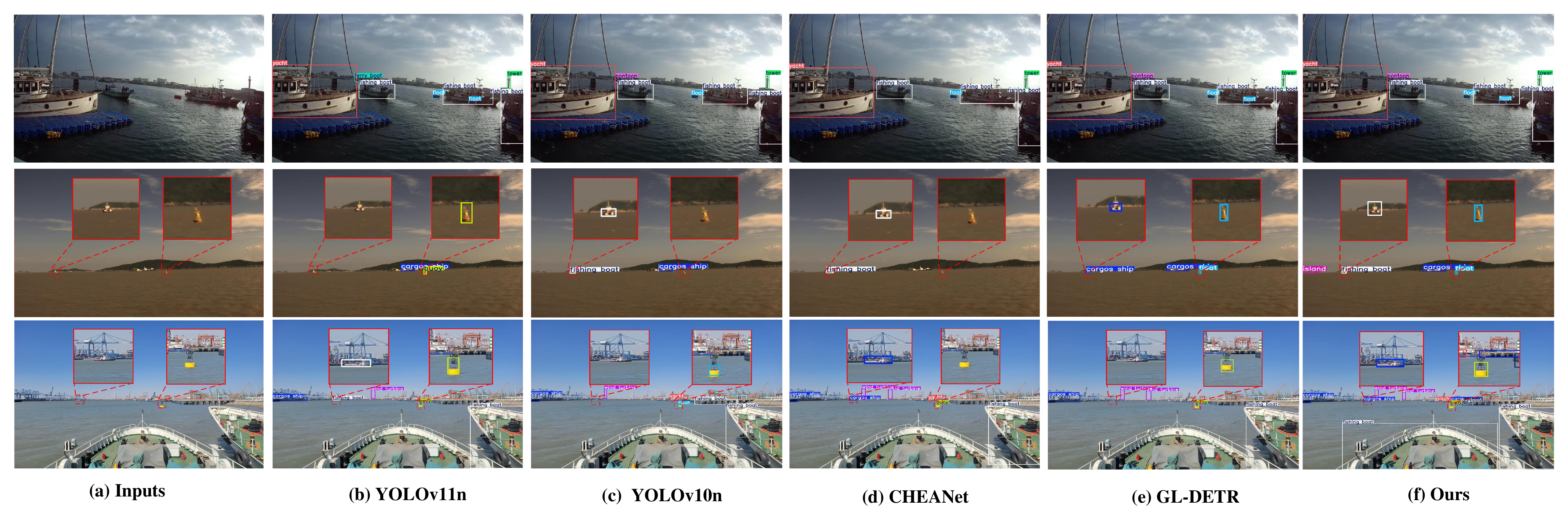}
\caption{Visualization Results of Different Detection Models show a comparison of our method with four state-of-the-art algorithms across three complex maritime scenarios—multi-scale multi-target, low-light at dusk, and wide depth of field—highlighting the superior performance of our method in these challenging conditions.}
\label{fig:5}
\end{figure*}
\subsection{Datasets}
This study presents the first dataset specifically designed for maritime target detection and channel planning from a shipborne perspective, Navigation12, to fill the gap in existing datasets in this domain. Our dataset consists of shipborne perspective images covering 12 types of maritime targets: cargo ships, transport ships, workboats, passenger ferries, speedboats, freighters, cruise ships, buoys, lighthouses, islands, wind turbines, and beacons. It is rich in data, with a total of 14,575 training images, 2,205 validation images, and 1,852 test images. Furthermore, the data collection process encompasses a variety of maritime regions, meteorological conditions, and multi-scale typical maritime scenarios, ensuring diversity and generalizability. This dataset provides a crucial foundation for multi-scale maritime target detection and intelligent navigation, and will effectively promote further research and applications in this field.

\subsection{Details and Evaluation Metrics}
We implemented the training and experiments of HMPNet using Python and PyTorch on an Ubuntu server equipped with an NVIDIA 4090 GPU. The optimizer used for model training was Stochastic Gradient Descent (SGD), with an initial learning rate set to 0.01, momentum parameter of 0.937, and weight decay of ${5e^{-4}}$. The batch size was set to 64, with a total of 200 epochs for training. The input image size was 640×640. To further enhance training performance, we employed a cosine annealing learning rate decay strategy to gradually reduce the learning rate. Additionally, to improve the generalization ability of the propsed model, we applied data augmentation techniques, such as rotation and flipping, to the training data.

To evaluate the performance of the proposed model, we used the mean Average Precision (mAP) metrics at IoU = 0.5 (mAP.50) and IoU = 0.5:0.05:0.95 (mAP.50:.95), along with the number of parameters (Params), giga floating-point operations per second (GFLOP), and frames per second (FPS). The Intersection over Union (IoU) measures the overlap between predicted and ground-truth bounding boxes, serving as a threshold to determine true positives in detection tasks. The mAP metrics assess detection accuracy, while Params and GFLOP indicate the model’s parameter size and computational complexity, and FPS reflects inference speed.

\subsection{Comparative Analysis}
\textbf{Quantitative Analysis.} We compared our proposed HMPNet with several mainstream object detection models, which is illustrated in Table \ref{tab:I}. They are classic ones like Faster R-CNN \cite{r12}, the YOLO series \cite{r14,r15,r16,r17,r18}, RetinaNet \cite{r27}, and some models specifically designed for maritime images, such as BLNet \cite{r6}, CHEANet \cite{r7} , and GL-DETR \cite{r5}. As presented in Table \ref{tab:I}, HMPNet demonstrates significant advantages in both accuracy and computational efficiency. It achieves 80.9$\%$ mAP@50, which is a 3.5$\%$ improvement over YOLOv10n and a 3.3$\%$ improvement over the latest SOTA model YOLOv11n. Compared to maritime object detection methods like GL-DETR and CHEANet, our method outperforms them by 3.5$\%$ and 3.2$\%$, respectively. This improvement is largely due to the HDM backbone, which efficiently performs multi-scale feature aggregation through dynamic feature modulation. Combined with the MCP neck, which enhances feature interaction and aggregation, it effectively reduces background noise and strengthens target recognition. Furthermore, HMPNet has only 2.0M parameters and 5.2G FLOPs, significantly lower than RT-DETR (29.3M, 105.2G) and CHEANet (18.6M, 102.2G). Compared to the current SOTA model (YOLOv11n), HMPNet reduces the number of parameters by approximately 0.6M, decreases computational complexity by 1.1G, and improves inference speed by 15.6 FPS. This improvement is primarily due to the efficient group convolutions of the MCP neck, which reduce computational resource consumption, and the PWS detector, which uses weight sharing mechanisms to reduce detection parameters while improving feature representation capabilities.

\begin{table}[!t]
\caption{\textbf{Comparison Experiments with SOTA Models.}} 
\label{tab:I}
\centering 
\setlength\tabcolsep{3pt}
\scalebox{0.9}{
\begin{tabular}{@{}ccccccc@{}}
\toprule 
Method & mAP.50 & mAP.50:95 & Params (M) & FLOPs (G) & FPS \\
\midrule 
Faster R-CNN \cite{r12} & 63.2 & 40.6 & 41.35 & 134 & 121.5 \\
YOLOv5n \cite{r15} & 78.2 & 56.4 & 2.2 & 5.8 & 343.1 \\
RetinaNet \cite{r27} & 69.3 & 46.9 & 36.33 & 128 & 159.1 \\
YOLOv6n \cite{r16} & 77.1 & 56.6 & 4.2 & 11.5 & 300.1 \\
RT-DETR \cite{r19} & 77.8 & 57.4 & 29.3 & 105.2 & 531.3 \\
YOLOv10n \cite{r17} & 78.2 & 56.3 & 2.3 & 6.5 & 369.7 \\
YOLOv11n \cite{r18}    & 78.3 & 56.5 & 2.6 & 6.3 & 376.0 \\
BLNet \cite{r6} & 77.5 & 57.2 & 47.8 & 146 & 210.5 \\
CHEANet \cite{r7} & 78.4 & 57.8 & 18.6 & 102.2 & 154.0 \\
GL-DETR \cite{r5} & 78.2 & 56.9 & 32.3 & 108.6 & 134.3 \\
LPEDet \cite{r4} & 76.3 & 53.5 & 5.68 & 18.4 & 263.1 \\
\midrule 
HMPNet (Ours) & \textbf{80.9} & \textbf{59.1} & \textbf{2.0} & \textbf{5.2} & \textbf{391.6} \\
\bottomrule 
\end{tabular}
}
\end{table}

\textbf{Qualitative Analysis.} Due to space constraints, we only present the results of the four best-performing models in comparison with our proposed method. These models include CHEANet and GL-DETR, which are highly effective in maritime object detection tasks, and YOLOv10n and YOLOv11n, which excel as general-purpose algorithms. The visualization results for three representative maritime scenarios, as shown in Figure \ref{fig:5}, include multi-scale multi-object scenes, low-light twilight scenes, and open-depth scenes. In the first row of Fig. \ref{fig:5}, it can be observed that in multi-scale multi-object scenes, YOLOv11n and CHEANet exhibit cases of missed and false detections for small objects, whereas HMPNet achieves accurate detection without such issues. This is primarily due to the HDM backbone and MCP neck, which effectively capture fine-grained details of small objects and enhance the aggregation of critical multi-scale information. The second row of Fig. \ref{fig:5} illustrates detection performance in low-light twilight scenes. The four other methods consistently exhibit missed detections, whereas HMPNet achieves superior detection performance by integrating a dynamic feature modulation mechanism. This mechanism markedly enhances object contrast, particularly under low-light conditions. The third row of Fig. \ref{fig:5} shows the detection outcomes in open-depth scenarios. It is evident that YOLOv10n and GL-DETR have considerable inaccuracies in bounding box annotations for distant objects. In contrast, HMPNet, through the synergistic operation of its MCP neck and PWS detector, achieves an excellent balance in detecting both near and distant objects.

The results demonstrate HMPNet’s robust adaptability in complex maritime scenarios. It surpasses other models in detection accuracy while optimizing parameters and computational efficiency, underscoring its potential for real-world intelligent navigation across diverse maritime environments.

\subsection{Ablation Experiments}
To validate the effectiveness of each module in HMPNet, we conducted ablation experiments on the proposed dataset, with results shown in Table \ref{tab:II}. When only the MCP neck was used, compared to the baseline model YOLOv11n, mAP.50 increased to 79.9$\%$. It indicates that the feature fusion layer effectively aggregates multi-scale features through matrix cascading multi-scale convolutions, significantly enhancing the adaptability of the model to targets of various scales. After replacing the detection head with the PWS detector, the number of parameters and FLOPs decreased by 15$\%$ and 4.7$\%$, respectively, while maintaining comparable mAP performance. It demonstrates its effective balance of accuracy and efficiency through shared convolution weights. Finally, after replacing the backbone with the HDM network, the model's mAP.50 and mAP.50:95 reached 80.9$\%$ and 59.1$\%$, respectively, achieving optimal performance. At the same time, the number of parameters and FLOPs decreased by 23$\%$ and 17$\%$, respectively. It reveals that the HDM backbone further optimized computational efficiency through hierarchical dynamic modulation. These experimental results validate the effectiveness of the HDM backbone, MCP neck, and PWS detector modules in improving both detection accuracy and computational efficiency in HMPNet.

\begin{table}[!t]
\caption{\textbf{Ablation Experiments of the Proposed HMPNet.}} 
\label{tab:II}
\centering 
\setlength\tabcolsep{3pt} 
\scalebox{0.9}{ 
\begin{tabular}{@{}ccccccc@{}}
\toprule 
MCPC & PWS & HDM & mAP.50 & mAP.50:95 & Params(M) & FLOPs (G) \\
\midrule 
$\times$ & $\times$ & $\times$ & 78.3 & 56.5 & 2.6 & 6.3 \\
$\checkmark$ & $\times$ & $\times$ & 79.9 & 58.7 & 2.5 & 6.3 \\
 $\checkmark$ & $\checkmark$ & $\times$ & 79.4 & 57.9 & 2.2 & 6.0 \\
$\checkmark$ & $\checkmark$ & $\checkmark$ & \textbf{80.9} &\textbf{ 59.1} & \textbf{2.0} & \textbf{5.2} \\
\bottomrule 
\end{tabular}
}
\end{table}
\section{Conclusion}

To promote the intelligent development of shipborne visual perception systems, this paper introduced the first-ever shipborne perspective object detection dataset, covering a variety of maritime areas and weather conditions and annotated with 12 different target classes. To the best of our knowledge, this is the first shipborne perspective object detection dataset. Leveraging this dataset, we propose HMPNet, an innovative architecture designed specifically for maritime multi-scale object detection, aimed at addressing challenges such as large variations in object scales, diverse target categories, and complex environmental interferences in maritime scenes. HMPNet achieves efficient cross-scale feature extraction, aggregation, and reuse through the design of the HDM backbone, MCP neck, and PWS detector, significantly enhancing both detection accuracy and efficiency. Experimental results demonstrate that HMPNet surpasses current mainstream detection models in terms of both accuracy and computational efficiency, with exemplary performance in small object detection and complex background handling. We hope that HMPNet will provide new insights for future research in intelligent ship navigation and offer robust technical support for practical applications such as maritime safety monitoring.

\bibliographystyle{IEEEbib}
\bibliography{icme2025references}

\vspace{12pt}

\end{document}